\documentclass[]{iflytek_robotics}

\usepackage{minitoc}
\usepackage{url}
\usepackage{amssymb}
\usepackage[utf8]{inputenc}
\usepackage{microtype}
\usepackage{booktabs}
\usepackage{pifont} 
\usepackage{multirow}
\usepackage{makecell}
\usepackage{paralist}
\usepackage{xspace}
\usepackage{color}
\usepackage{xcolor}
\usepackage{colortbl}
\usepackage{adjustbox}
\usepackage{hyperref} 
\usepackage[edges]{forest}
\usepackage{tikz} 
\usepackage{caption}
\usepackage{amsfonts}
\usepackage[toc,page,header]{appendix}
\usepackage[utf8]{inputenc}
\usepackage{bbm}
\usepackage{enumitem}

\definecolor{seedc}{RGB}{7, 92, 173}

\newcommand{\name}[1]{iFlyBot-VLA}

\renewcommand{\paragraph}[1]{\vspace{0.1em}\noindent\textbf{#1}}

\title{iFlyBot-VLA Technical Report}

\author[1,\dag]{Yuan Zhang} 
\author[1,\dag]{Chenyu Xue}
\author[1,\dag]{Wenjie Xu}
\author[2]{Chao Ji}
\author[1]{Jiajia wu}
\author[1,*]{Jia Pan}

\affiliation[1]{iFlyTek Reasearch and Development Group}
\affiliation[2]{LindenBot}

\abstract{
We introduce \textbf{\name{}}, a large-scale Vision-Language-Action (VLA) model trained under a novel framework.
The main contributions are listed as follows:
(1) a \textbf{latent action model} thoroughly trained on large-scale human and robotic manipulation videos;
(2) a \textbf{dual-level action representation framework} that jointly supervises both the Vision-Language Model (VLM) and the action expert during training;
(3) a \textbf{mixed training strategy} that combines robot trajectory data with general QA and spatial QA datasets, effectively enhancing the 3D perceptual and reasoning capabilities of the VLM backbone.
Specifically, the VLM is trained to predict two complementary forms of actions:
\textbf{latent actions}, derived from our latent action model pretrained on cross-embodiment manipulation data, which capture implicit high-level intentions; and
\textbf{structured discrete action tokens}, obtained through frequency-domain transformations of continuous control signals, which encode explicit low-level dynamics.
This dual supervision aligns the representation spaces of language, vision, and action, enabling the VLM to directly contribute to action generation.
Experimental results on the \textbf{LIBERO Franka} benchmark demonstrate the superiority of our framework, while real-world evaluations further show that \name{} achieves competitive success rates across diverse and challenging manipulation tasks.
Furthermore, we plan to \textbf{open-source a portion of our self-constructed dataset} to support future research in the community.

}

\date{\today}
\correspondence{\email{chaoji@iflytek.com, yuanzhang10@iflytek.com}}
\checkdata[Project Page]{\url{https://xuwenjie401.github.io/iFlyBot-VLA.github.io/}}

\begin{document}
\maketitle

\section{Introduction}
\label{sect:intro}

In recent years, with the rapid advancement of Vision-Language Models (VLMs)\cite{achiam2023gpt,bai2025qwen2}, researchers have begun to envision leveraging their powerful perceptual and reasoning capabilities to accomplish long-horizon and complex robotic manipulation tasks.
However, while the autoregressive paradigm has demonstrated impressive performance in complex scene understanding and perception, it faces inherent limitations when dealing with tasks that require precise, continuous control signals—such as predicting joint angles or end-effector poses. Compared with continuous modeling approaches like diffusion or flow-based methods, autoregressive models struggle to generate fine-grained, numerically accurate outputs.

As a result, most contemporary Vision-Language-Action (VLA) frameworks\cite{zitkovich2023rt,kim2024openvla,team2024octo,chen2025internvla,black2024pi0,intelligence2025pi_,cheang2025gr,team2025gemini} adopt a hybrid design that integrates a VLM\cite{beyer2024paligemma,chen2023pali} for perception and a diffusion or flow-based action expert for motion generation. This combination provides strong input–output compatibility and leverages the complementary strengths of both components. Nonetheless, a key challenge remains: how to design a training strategy that maximally preserves the VLM’s general perception and reasoning capabilities while enabling the diffusion policy to produce precise, smooth actions.

On the other hand, imitation learning remains the foundation of most current VLA systems, which rely heavily on high-quality teleoperation datasets \cite{jiang2011efficient,pinto2016supersizing,kappler2015leveraging,mahler2017dex,depierre2018jacquard,kalashnikov2018scalable,eppner2021acronym,zhu2023fanuc,khazatsky2024droid,wu2024robomind,jiang2025galaxea} collected via specialized interfaces. Although recent works have explored alternative data sources such as VR teleoperation \cite{arunachalam2022holo,george2025openvr,tsokalo2019remote,cheng2024open,he2024omnih2o}, handheld devices \cite{chi2024universal}, and human demonstration datasets \cite{damen2018scaling,goyal2017something,grauman2022ego4d,wang2023holoassist,hoque2025egodex,liu2022hoi4d,wang2024egovid}, teleoperated data with consistent morphologies still holds distinct advantages in precision and consistency.
While large-scale and diverse manipulation data are crucial for improving generalization across tasks and embodiments, one motivation for integrating VLMs into VLAs is to leverage their broad perceptual and semantic generalization. However, when trained solely on manipulation data, the VLM’s language and reasoning abilities degrade rapidly. Thus, an open challenge remains: how to balance multimodal training so that both the VLM and the action expert can co-evolve—maintaining perception, language understanding, and action generation throughout training.

Early Vision-Language-Action (VLA) research explored learning manipulation policies via autoregressive Vision-Language Models (VLMs), facing the challenge of discretizing continuous action sequences. OpenVLA \cite{kim2024openvla} normalized continuous actions into 256 discrete bins, but suffered from precision and scalability issues as action chunks grew. FAST \cite{black2025real} addressed this by applying a Discrete Cosine Transform (DCT) \cite{ahmed2006discrete}–based compression followed by Byte Pair Encoding (BPE) \cite{shibata1999byte}, significantly improving learning efficiency and action accuracy.

Recent approaches learn manipulation knowledge from large-scale videos through latent action representations. LAPA \cite{ye2024latent} uses a VQ-VAE \cite{van2017neural} to discretize latent action increments in unlabeled videos, training a VLA to predict these and fine-tuning on limited robot data for real-world control. UniVLA \cite{bu2025univla} extends this with a two-stage task-centric framework, introducing task-agnostic and task-specific codebooks for better noise filtering and cross-domain generalization. These studies demonstrate that discretized latent action spaces effectively bridge visual–language perception and fine-grained action generation for scalable manipulation learning.

In this report, we present \textbf{iFlyBot-VLA}, a framework trained in multiple stages using diverse types of data. 
\textbf{iFlyBot-VLA} takes natural language task instructions, multi-view RGB images of the environment, and the robot’s proprioceptive state data as inputs, and outputs \textit{action chunks} to control a dual-arm robot in an end-to-end manner. 
Specifically, iFlyBot-VLA builds upon a pretrained Vision-Language Model (VLM) and predicts action chunks through a \textbf{flow-matching} mechanism.

We observe that a randomly initialized flow-based action expert, when trained end-to-end solely on robot trajectory data, can easily degrade the general perceptual ability of the VLM backbone. 
However, this backbone capability is crucial for the policy’s generalization performance. 



To address the aforementioned challenges, we have made the following contributions:
\begin{itemize}
    \item We collect and organize a large-scale, high-quality dataset consisting of single-arm, dual-arm, and human teleoperation videos to train a \textbf{latent action model}, thereby obtaining high-level and generalizable latent action representations.
    
    \item Based on the latent action representations and explicit supervision of structured discrete actions, we construct a \textbf{dual-level action representation framework}, enabling the joint training of the VLM and the action expert.
    
    \item To preserve the VLM’s general understanding capability and further enhance its embodiment-related spatial perception, we build extensive \textbf{general vision-language QA} and \textbf{spatial reasoning QA} datasets. These are carefully mixed with robot manipulation data using an optimized ratio, improving the policy’s generalization performance.
    
    \item We design comprehensive comparative experiments on both the \textbf{LIBERO benchmark} and \textbf{real-world robotic platforms}. The results demonstrate that our proposed framework achieves superior performance across diverse tasks in both simulation and real environments.
    
    \item We will \textbf{open-source} our code, along with portions of our teleoperation and VQA datasets, to contribute to the research community.
\end{itemize}

\section{Overview}
\label{sect:overview}

\begin{figure}[ht]
    \centering
    \includegraphics[width=\textwidth]{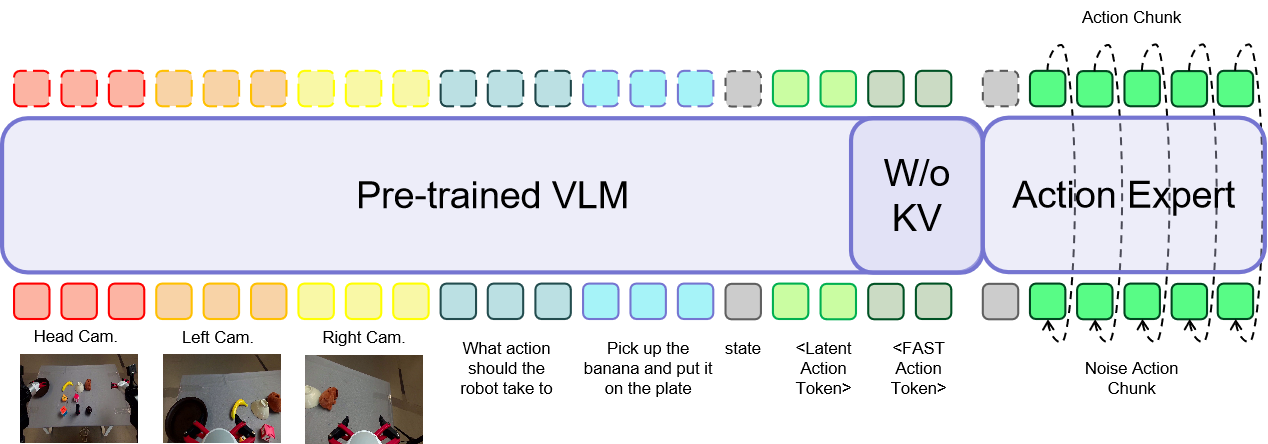} 
    \caption{The architecture of iFlyBot-VLA consists primarily of a language transformer backbone and an action expert network. The model generates executable robot actions through a combination of explicit and implicit planning. The key–value (KV) cache from the VLM component is passed to the downstream action expert, while the FAST Action Token—which corresponds to the implicit planning process—is not forwarded to the Action Expert} 
    \label{fig:model} 
\end{figure}

Figure \ref{fig:model} provides an overview of the \name{} model and its overall training pipeline. In this framework, we first construct a manipulation dataset of robot trajectories, which is a weighted combination of three major components: self-collected teleoperation dataset by iFLYTEK, a subset of OXE dataset \cite{o2024open},and a subset of AgiBot-World dataset \cite{bu2025agibot}.

During the pre-training phase, \name{} also leverages a portion of pure text-based question-answering data to enhance the model’s spatial reasoning and object perception capabilities. This text-only dataset is internally constructed by iFLYTEK and focuses on spatial understanding tasks.

As mentioned in the Introduction, in order to learn high-level latent action representations from a broader range of manipulation videos—thereby benefiting the learning of downstream numerical action chunks—we first train a more comprehensive codebook via the VQ-VAE pipeline before initiating the VLA training. This codebook serves as the foundation for providing latent action tokens that guide subsequent learning.
Our overall training framework is divided into three stages:

1) \textbf{Stage I: Latent Action Training.}  
We first train a \textbf{latent action model} using large-scale, carefully curated human and robotic manipulation videos. Through a VQ-VAE–based architecture, the model learns to extract high-level and generalizable latent action representations from pairs of consecutive frames, providing rich supervision signals that benefit subsequent VLA training.

2) \textbf{Stage II: Foundational Pre-training.}  
The goal of this stage is to build a foundational model with broad spatial perception, object recognition, and generalization capabilities, without specializing in any particular downstream task. The resulting model can follow natural-language commands, execute instruction-driven actions, and recognize both object categories and their spatial relations.

3) \textbf{Stage III: Task-specific Post-training.}  
For more complex and dexterous operations, we conduct a second-stage post-training process using high-quality, self-collected datasets to adapt the model to specific downstream tasks. \name{} employs large-scale, high-fidelity post-training strategies for sophisticated robotic skills such as cloth folding, manipulation in cluttered scenes, and grasp-and-place tasks involving irregular objects. This stage relies on larger datasets to achieve high-precision control and robust adaptability.



The next chapter provides a detailed explanation of the \name{} model architecture. The model builds upon the Qwen2.5-VL (3B) \cite{bai2025qwen2} vision-language backbone. By further training on the hybrid dataset—comprising both spatial question-answering data and robot manipulation data—the vision-language model is enhanced to understand and generate action-related responses.

To transform Qwen2.5-VL into a vision-language-action (VLA) model capable of generating continuous action trajectories, a downstream Action Expert Module is introduced. This module is based on Flow-Matching Diffusion Transformer \cite{peebles2023scalable} architecture, which performs denoising via flow-matching techniques using the KV-features extracted from the vision-language model. The design and implementation of this component are described in detail in the following section.

It is worth noting that \name{} adopts Qwen2.5-VL primarily due to its superior performance and deployment flexibility. In principle, the same framework can be adapted to any vision-language backbone with minimal modification.
\section{\name{} Model}
\label{sect:model}

The architecture of \name{} consists primarily of a language transformer backbone and an action expert network.
The model generates executable robot actions through a combination of explicit and implicit planning.
The key–value (KV) cache from the VLM component is passed to the downstream action expert, while the FAST Action Token—which corresponds to the implicit planning process—is not forwarded to the Action Expert.

The \textbf{\name{}} model is an end-to-end \textbf{Vision-Language-Action (VLA)} model denoted as $\pi_{\theta}$. It controls a dual-arm robot by generating an \textbf{action block} of length $k$, represented as $a_t = a_{t:t+k}$. The generation of this action block primarily depends on three inputs: the language instruction $l$, the visual observation $o_t$ obtained from sensors, and the current robot state $s_t$, formulated as \ref{eqn:model}:

\begin{equation}
a_t = \pi_{\theta}(l, o_t, s_t)
    \label{eqn:model}
\end{equation}

The model mainly consists of a language transformer backbone and an action expert network, which together generate executable robot actions $a_t$ through explicit and implicit planning. The supervision for latent action tokens is provided by an expert network based on the visual changes across future frames, while the discretized action tokens are obtained by an action token encoder that encodes a sliding window of future actions. In practice, we assign these action tokens to the unused token IDs within the VLM tokenizer to ensure seamless integration with the language modeling framework.

\subsection{latent action model}



\begin{figure}[ht]
    \centering
    \includegraphics[width=\textwidth]{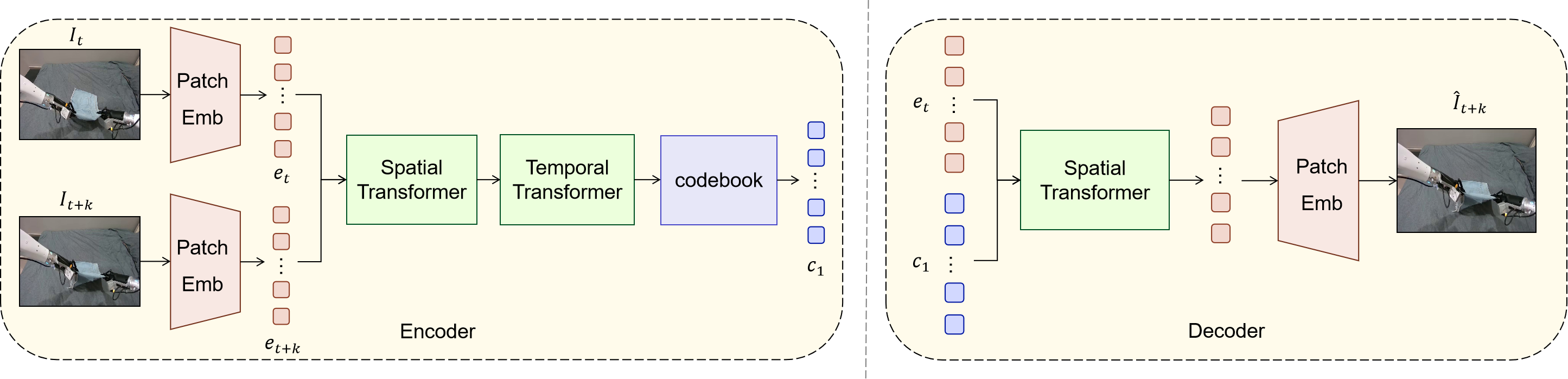} 
    \caption{Architecture of the latent action token encoding expert network} 
    \label{fig:lam} 
\end{figure}

To learn latent actions in a self-supervised manner, we train a \textbf{latent action model} with an encoder–decoder structure, which serves as the expert network, as shown in Fig.~\ref{fig:lam}.

The encoder consists of both spatial and temporal Transformers, taking as input a set of current image frames $o_t$ and future frames $o_{t+k}$ spaced by $k$, and outputs the latent action $c_t$. 
To ensure sufficient motion variation between frames, the interval between the two frames is fixed at 1 second, and the corresponding frame gap $k$ is determined according to the frame rate of each dataset. 

The decoder, which contains only a spatial Transformer, takes the latent action $c_t$ and the image frame $o_t$ as inputs to reconstruct the future frame $o_{t+H}$.

The latent action quantization model is trained based on the \textbf{VQ-VAE objective}, discretizing the continuous encoded features by retrieving the nearest quantized representation from the codebook, which facilitates the VLM’s learning of $z_t$ by \ref{eqn:LAPA1}:
\begin{equation}
    c_t = \arg\min_n \|x_{\text{enc}} - c_n\|^2
    \label{eqn:LAPA1}
\end{equation}
The size of the codebook $|C|$ is set to 32, and 8 discrete codes are retrieved at each step. 
During reproduction of the VQ-VAE training, we encountered a gradient collapse issue. 
To address this, we adopt the \textbf{NSVQ algorithm}~\cite{vali2022NSVQ}, which replaces the original VQ’s Straight-Through Estimator (STE) with a noise-based approximation. 
Specifically, it substitutes the quantization error between the discrete output and the encoder output with the product of this error and a unit noise vector, thereby preventing gradient collapse by \ref{eqn:LAPA2}:
\begin{equation}
    c_t' = x_{\text{enc}} + \frac{\|x_{\text{enc}} - c_t\|}{\|w\|} w
    \label{eqn:LAPA2}
\end{equation}
where $w$ follows a standard Gaussian distribution.

Additionally, during the decoding process, we apply a \textbf{stop-gradient} operation to the current frame, forcing the model to rely on the latent action $c_t$ for decoding the future frame. 
The supervision loss is defined as the mean squared error (MSE) between the reconstructed and target images.

\subsection{Discrete Action Token Encoding}

The discrete action tokens in the iFlyBot-VLA model are encoded by an \textit{action token encoder} based on the action window $a_t$ composed of multiple future action frames. Specifically, we adopt the \textit{Fast Action Token} method~\cite{black2025real} to encode $a_t$. 

It is worth noting that, on one hand, the discrete action tokens share a high degree of similarity with the downstream action outputs generated by the action expert, which may lead to overfitting if their corresponding features are directly provided to the expert, thereby reducing the model’s generalization ability. On the other hand, the number of discrete action tokens is relatively large, which can significantly increase inference time and reduce efficiency when generating them in real-time. 

Therefore, in this work, the discrete action tokens are used \textbf{only to supervise the VLM (Vision-Language Model) component} to implicitly encourage the model to learn action-related semantics and to assist in implicit action planning. During both training and inference, the corresponding features of the discrete action tokens are not utilized. In contrast, the \textbf{latent action tokens} are constructed from a more compact latent action space, whose features are used for downstream action planning. This design enhances the quality of robot action generation while maintaining high inference efficiency.

\subsection{VLA Model Architecture}

The iFlyBot-VLA model $\pi_{\theta}$ consists of two main components: 
(1) a \textbf{Transformer-based vision-language backbone} for embedding multimodal states, and 
(2) a \textbf{Diffusion Transformer expert network} for generating continuous robot actions.

In the language Transformer backbone, we adopt a standard \textit{decoder-only} vision-language model (VLM) framework. The image encoder embeds visual observations from the robot’s onboard cameras into the same feature space as the language tokens. In addition to conventional vision-language inputs, iFlyBot-VLA introduces \textbf{robot-specific inputs}, namely the robot’s proprioceptive state. In practice, we insert a placeholder token for the state input, which is later replaced by the feature obtained from passing the actual state values through a fully connected layer. During training, the VLM is supervised to predict two sequences: a set of \textit{latent action tokens} and a set of \textit{discrete action tokens}. The latent action tokens assist in the planning and generation of the action block $a_t$, while the discrete tokens guide the VLM in learning spatially grounded, action-related semantics.

The downstream action expert network is a \textbf{Diffusion Transformer}. During inference, the VLM outputs KV caches at every layer, which are fed into the action expert to provide necessary visual-linguistic context and planning information. Only the KV caches corresponding to the \textit{latent action tokens} are retained, while those related to the discrete action tokens are excluded. The KV caches of latent action tokens are preserved because they provide a highly compressed representation of actions, effectively supporting downstream planning. In contrast, discrete action tokens are excluded since their large quantity would slow down inference and reduce overall efficiency.

For modeling continuous actions, iFlyBot-VLA adopts the \textbf{flow-matching} approach~\cite{liu2022rectified,lipman2022flow} to represent continuous action distributions. This method is particularly suitable for high-frequency manipulation tasks. In implementation, the downstream expert network takes as input the robot state, the current timestep, and a \textbf{noised action} composed of the target action window, timestep, and Gaussian noise. Together with the expert’s features, the model predicts the denoising direction from the noised action toward the true action. The corresponding loss is defined as:
\begin{equation}
    L^{\tau}(\theta) = \mathbb{E}_{p(A_t|l,o_t,s_t),q(A_t^\tau|A_t)} \left\| \pi_\theta(A_t^\tau,l,o_t,s_t) - \pi(A_t^\tau|A_t) \right\|^2
    \label{eqn:loss}
\end{equation}

Here, $t$ denotes the robot’s timestep and $\tau \in [0,1]$ represents the flow-matching time variable. During training, Gaussian noise $\epsilon \sim \mathcal{N}(0,1)$ is sampled, and the noised action is constructed as $A_t^\tau = \tau A_t + (1-\tau)\epsilon$. The network output $\pi_\theta(A_t^\tau,l,o_t,s_t)$ is trained to match the denoising vector field $\pi(A_t^\tau|A_t) = \epsilon - A_t$. The action expert module adopts a \textbf{fully bidirectional attention mask}, ensuring that all action tokens within the same window can attend to each other, allowing parallel denoising across the action window. This design both promotes temporal continuity between actions and improves generation efficiency. During training, the flow-matching timestep $\tau$ is sampled from a Beta distribution that assigns higher weights to smaller (noisier) timesteps.

\subsection{Inference Process}

During inference, iFlyBot-VLA generates actions starting from random Gaussian noise $A_t^0 \sim \mathcal{N}(0,1)$ and integrates the learned vector field from $\tau = 0$ to $\tau = 1$ to obtain the final action sequence. Specifically, a discrete forward Euler integration is performed as \ref{eqn:val}:
\begin{equation}
    A_t^{\tau+\sigma} = A_t^{\tau} + \sigma \pi_\theta(A_t^\tau, l, o_t, s_t)
    \label{eqn:val}
\end{equation}
where $\sigma$ is the integration step size. In practice, a five-step integration is used, i.e., $\sigma = 0.2$. During inference, the VLM’s KV cache needs to be computed only once, regardless of the number of integration steps, ensuring efficient and stable action generation.

\section{Training Strategy}
\label{sect:strategy}

\subsection{Data Preparation}

Our latent action network does not rely on textual inputs, allowing for more diverse data usage. On one hand, we incorporate large-scale human manipulation datasets. The inclusion of these datasets enhances the generalization ability of the model, enabling it to handle complex manipulation scenarios and diverse tasks. On the other hand, the latent action network $F$ is trained using both single-arm and dual-arm robotic datasets. These datasets correspond to various robotic operation settings and motion types, enabling the expert model to effectively capture action dynamics across both single-arm and dual-arm robotic contexts.

\begin{figure}[ht]
    \centering
    \includegraphics[width=\textwidth]{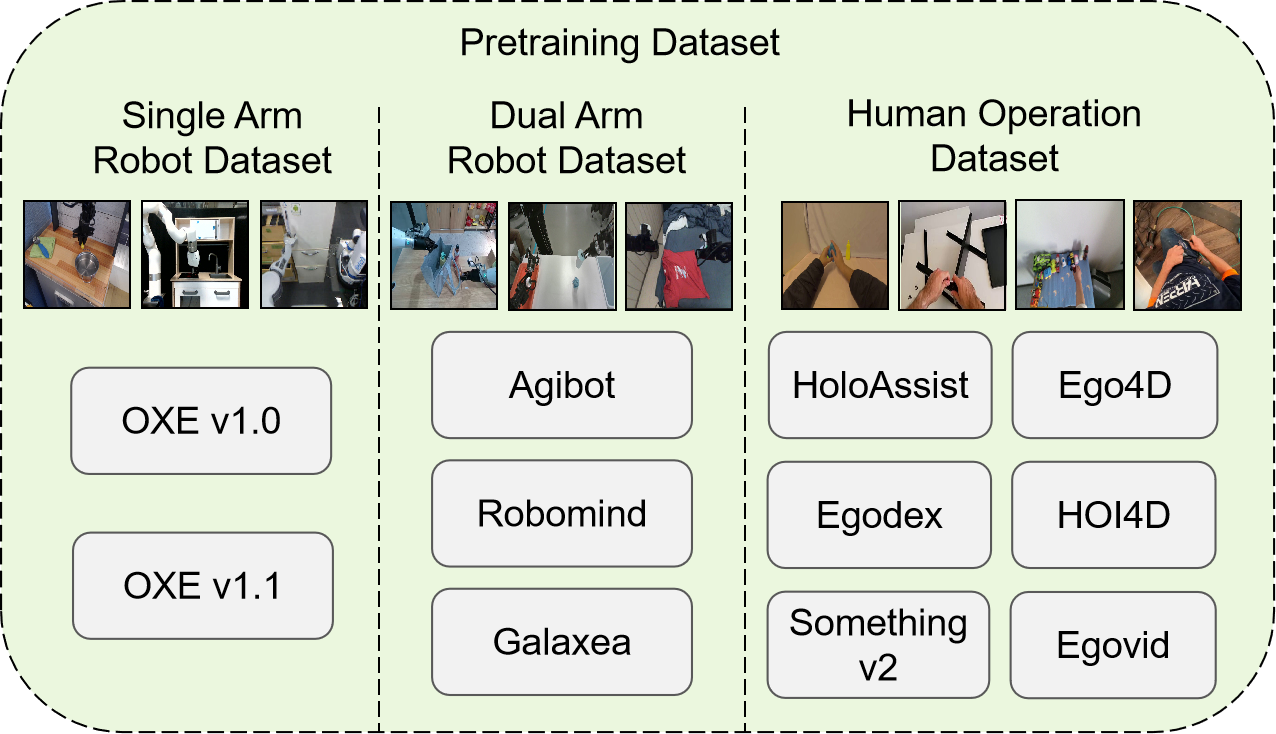} 
    \caption{The data used for training the latent action token encoding expert network} 
    \label{fig:lamdata} 
\end{figure}

Our latent action network does incorporate large-scale human manipulation datasets including HoloAssist~\cite{wang2023holoassist}, Ego4D~\cite{grauman2022ego4d}, EgoDex~\cite{hoque2025egodex}, HOI4D~\cite{liu2022hoi4d}, Something-Something V2~\cite{goyal2017something}, and EgoVid~\cite{wang2024egovid}, and using both single-arm and dual-arm robotic datasets, including OXE~\cite{o2024open}, AgiBot~\cite{bu2025agibot}, RoboMind~\cite{wu2024robomind}, and Galaxea~\cite{jiang2025galaxea}, as show in fig \ref{fig:lamdata}.

During the training of \name{}, the datasets used include: (1) an internally constructed \textbf{VQA dataset focuses on spatial understanding} developed by iFLYTEK, (2) a subset of the publicly available \textbf{OXE}\cite{o2024open} dataset containing a wide range of robotic manipulation tasks under diverse scenarios and robot embodiments, (3) a subset of the publicly released \textbf{AgiBot-World}\cite{bu2025agibot} dataset from AgiBot featuring dual-arm robotic operations, and (4) a \textbf{self-collected dataset} from iFLYTEK, which includes various dual-arm manipulation tasks such as tabletop pick-and-place, cloth folding, and grasping irregular or soft objects across multiple real-world environments.
\begin{figure}[ht]
    \centering
    \includegraphics[width=\textwidth]{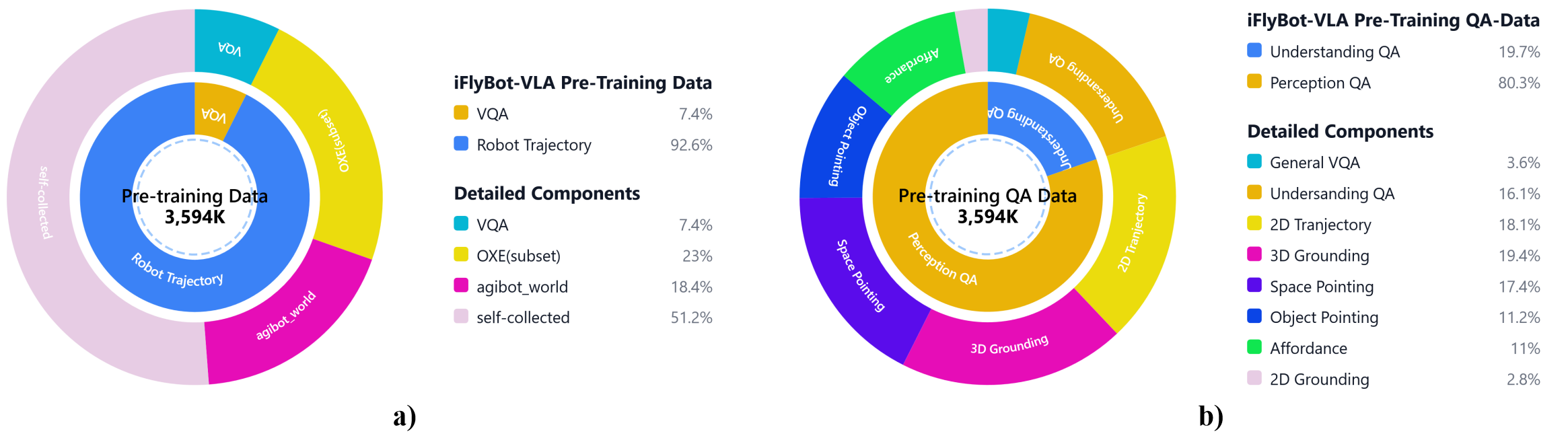}
    \caption{
        \textbf{Overview of our dataset.}
        The pretraining mixture consists of subsets of OXE, AgiBot\_World, self-collected manipulation data, and VQA data.
        The left figure shows the proportion of different datasets in the pretraining mixture,
        while the right figure illustrates the composition of QA datasets during the pretraining stage.
    }
    \label{fig:pretrain_data}
\end{figure}

The self-collected iFLYTEK dataset was gathered using \textbf{26 dual-arm robots} configured in two different setups. It contains three primary categories of tasks:
\begin{itemize}
    \item \textbf{Cloth Folding:} Includes five types of T-shirts and three types of shorts. For each type, 190 trajectories were collected, each averaging 4.5 minutes, totaling approximately 110 hours of data.
    \item \textbf{General Pick and Place:} Includes 30 categories of grasped objects, with 400 trajectories per category, each averaging 27 seconds, totaling around 90 hours of data.
    \item \textbf{Long-Horizon Parcel Sorting}: The goal is to handle flexible packages of varying sizes and flip them so that the label side faces upward. A total of 2,752 trajectories were collected, each averaging 61 seconds, resulting in approximately 47 hours of data.
\end{itemize}
Both the folding and grasping datasets are planned to be released publicly in the future.

The training of \name{} consists of two stages. In the \textbf{first pre-training stage}, we use the full text-based spatial QA dataset, the OXE and AgiBot open-source datasets, and the complete self-collected dual-arm manipulation dataset from iFLYTEK, which covers diverse scenes and tasks. In the \textbf{second fine-tuning stage}, the model is trained on task-specific datasets to adapt to particular robot actions. The data distribution ratio for pre-training is illustrated in Fig.~\ref{fig:pretrain_data}.

\subsection{Training Strategy}

In the \textbf{first-stage pre-training}, the full dataset—including the general spatial QA data—is utilized. For the QA data, \name{} employs special marker tokens. All action outputs for these samples are set to zero, and the action loss is not computed. Additionally, no gradient is propagated through the action expert module for these samples.

For robotic manipulation data, since the action expert is randomly initialized, the end-to-end training gradients will cause significant interference to the pre-trained VLM. This interferes with the preservation and learning of the VLM's capabilities. \name{} truncates the gradient flow from the action expert to the vision-language backbone.


In the \textbf{second-stage fine-tuning}, after pre-training, the vision-language backbone already possesses strong possesses strong capabilities in embodied perception, reasoning, and implicit action feature extraction. The focus of fine-tuning is therefore on enabling the action expert to learn robot-specific control dynamics and foster better interaction between upstream and downstream modules. During this stage, gradient propagation from the action expert to the vision-language backbone is enabled, because the gradients of the action experts in the randomly initialized Diffusion Transformer structure will undermine the capabilities of the pre-trained VLM component. Moreover, while the pre-trained VLM already possesses strong capabilities in embodied perception, reasoning, and implicit action feature extraction, the action expert still needs further adaptation to downstream tasks, \name{} for each batch of actions applies \textbf{multi-sample noise perturbations}—sampling different noisy versions of the same action sequence—to perform denoising and backpropagation jointly. This strategy accelerates training and enhances the stability of the action expert.

Throughout all training stages, \name{} pads both action and state vectors to a \textbf{20-dimensional space}, where the first 10 dimensions correspond to the left arm and the latter 10 to the right arm. For single-arm datasets, during the first-stage training, the action data are randomly assigned to either the left or right arm to maintain consistency in input dimensionality.

\section{Experiments}
\label{sect:experiments}

To systematically evaluate the performance of iFlyBot-VLA and the effectiveness of its individual components, we conducted experiments in both simulated and real-world environments. The iFlyBot-VLA model was first tested in the \textbf{LIBERO simulator}~\cite{liu2023libero}, where it was compared with several existing methods to validate its performance. In addition, ablation studies were performed within LIBERO to examine the respective contributions of explicit and implicit planning to the overall model performance. 

In real-world scenarios, iFlyBot-VLA was tested across three distinct task settings: (1) complex tabletop pick-and-place, (2) manipulation of irregularly shaped and deformable objects, and (3) cloth folding. The results demonstrate that iFlyBot-VLA achieves strong stability and generalization across diverse and challenging tasks, effectively learning robust policies capable of handling long-horizon and fine-grained manipulation.

\subsection{Results in the LIBERO Simulator}

\begin{figure}[ht]
    \centering
    \includegraphics[width=\textwidth]{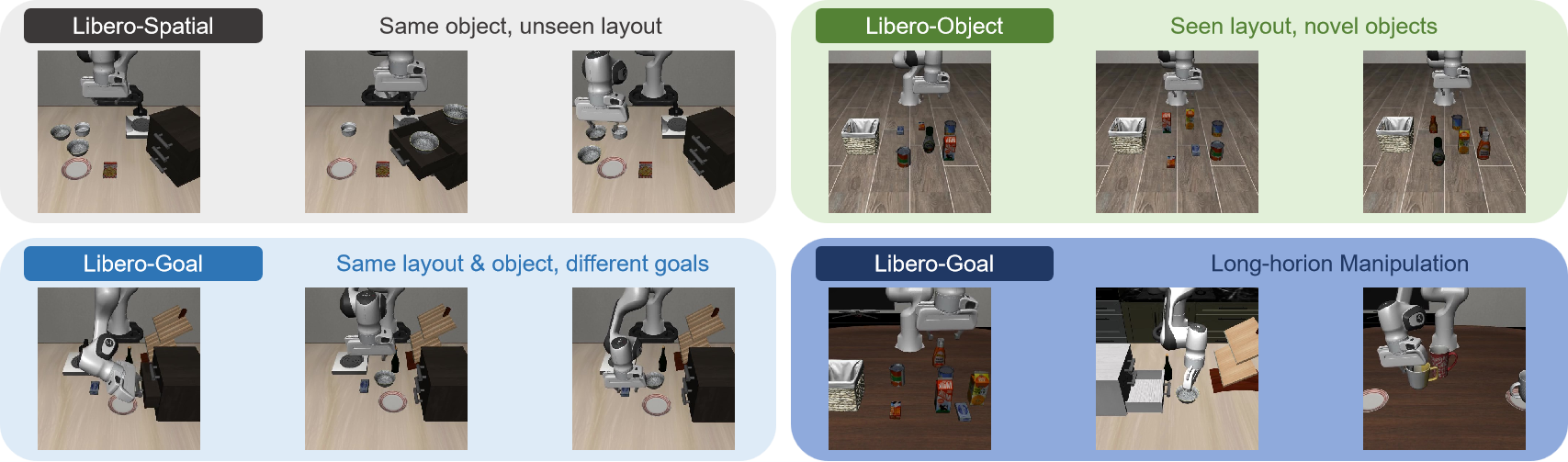}
    \caption{Task suites in the LIBERO dataset.}
    \label{fig:libero}
\end{figure}

The \textbf{LIBERO benchmark}~(see Fig.~\ref{fig:libero}) consists of four task suites designed to evaluate learning approaches in robotic manipulation. In this work, we focus on \textbf{supervised fine-tuning within target task suites}, using imitation learning combined with \textit{flow-matching denoising} to train policies on successful demonstration trajectories, and then assess their performance.

As illustrated in Fig.~\ref{fig:libero}, LIBERO includes four task suites, each containing 10 tasks with 10 demonstrations per task:
\begin{enumerate}
    \item \textbf{LIBERO-Spatial:} Tasks require spatial reasoning, such as accurately placing a bowl based on inferred geometric relationships, evaluating the model’s ability in spatial and geometric reasoning.
    \item \textbf{LIBERO-Object:} The scene layout remains fixed while object categories vary, testing the model’s generalization to new object instances.
    \item \textbf{LIBERO-Goal:} Object and scene layouts are fixed, but task goals differ, evaluating goal-directed adaptability and reasoning.
    \item \textbf{LIBERO-Long:} Long-horizon tasks composed of multiple sub-goals, diverse objects, layouts, and task sequences, testing the model’s ability to plan and execute multi-step manipulation.
\end{enumerate}

During training, we follow the \textbf{OpenVLA} data preprocessing pipeline by removing all failed demonstrations. iFlyBot-VLA was trained for \textbf{70,000 steps} on the LIBERO-Long suite and \textbf{50,000 steps} on the remaining suites, with a global batch size of 64 and an action window size of 7 for both training and inference. Only third-person camera images and textual task instructions were used as inputs. Importantly, none of the LIBERO samples overlapped with those used in pre-training or latent action model training, ensuring a stringent test of the model’s generalization ability.

The comparison baselines include three representative models, among which \textbf{OpenVLA} and \textbf{$\pi_0$} are most closely related to our method:
\begin{itemize}
    \item \textbf{LAPA}~\cite{ye2024latent}: An unsupervised framework that learns latent actions from unlabelled human videos.
    \item \textbf{OpenVLA}~\cite{kim2024openvla}: A vision-language-action model trained on large-scale and diverse datasets such as OpenX to achieve general-purpose robot policy learning.
    \item \textbf{$\pi_0$}~\cite{black2024pi0}: A vision-language-action model that generates continuous actions via an action expert trained on OpenX and self-collected robot data, then fine-tuned for downstream tasks.
\end{itemize}

\begin{figure}[ht]
    \centering
    \includegraphics[width=\textwidth]{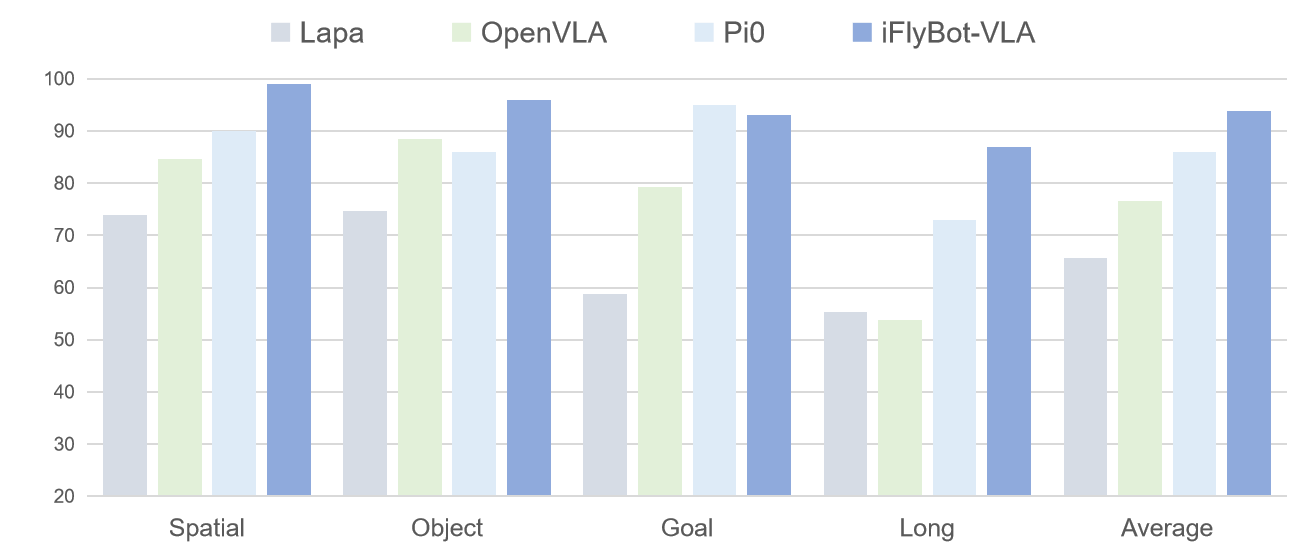}
    \caption{Comparison of iFlyBot-VLA and representative VLA models on the LIBERO simulator dataset.}
    \label{fig:compler}
\end{figure}

The experimental results are shown in Fig.~\ref{fig:compler}.

In this experiment, iFlyBot-VLA achieved an \textbf{average accuracy of 93.8\%} across LIBERO tasks, outperforming the best existing VLA model, $\pi_0$ (86\%), and significantly surpassing OpenVLA (76.5\%). iFlyBot-VLA achieved state-of-the-art performance in all task suites except LIBERO-Goal, where it still reached 93\%, nearly matching $\pi_0$’s 95\%. These results demonstrate that iFlyBot-VLA achieves strong and consistent performance within the LIBERO simulator, with clear improvements over previous action-window-based approaches.

\subsection{Ablation Study in LIBERO}

\begin{figure}[ht]
    \centering
    \includegraphics[width=\textwidth]{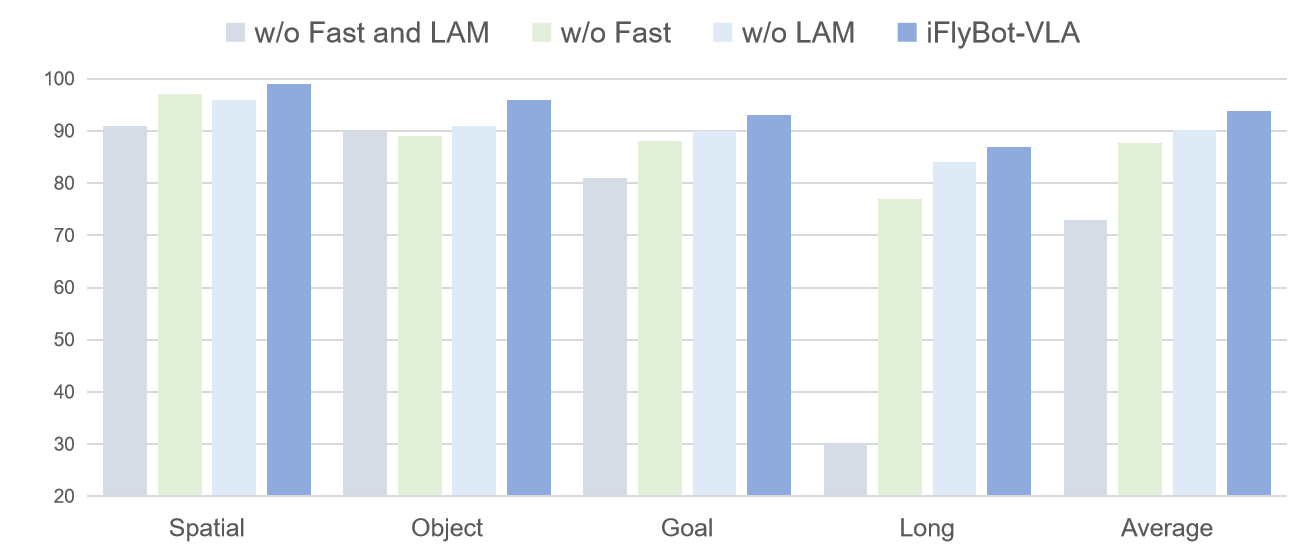}
    \caption{Ablation results showing the effect of different components of iFlyBot-VLA in the LIBERO simulator.}
    \label{fig:abla}
\end{figure}

To further investigate the roles of explicit and implicit planning, an ablation study was conducted in the LIBERO simulator. The results are presented in Fig.~\ref{fig:abla}.

The full iFlyBot-VLA model achieved the best overall performance. Compared to the version without the Fast module (\textit{w/o Fast}), which achieved 87.8\% success, the full model improved by 6\%. Compared to the version without the LAM module (\textit{w/o LAM}), which achieved 90.3\%, the full model improved by 3.5\%. When both Fast and LAM modules were removed (\textit{w/o Fast and LAM}), success dropped to 73\%, yielding a 20.8\% improvement when both were included. These results clearly demonstrate that both explicit and implicit planning mechanisms contribute positively to task execution, particularly for long-horizon manipulation tasks, where their combined effect is most pronounced.

\subsection{Real-World Experiments}
\label{sec:exp:ppa}

Our real-world experiments consist of three main parts:
\begin{itemize}[left=0pt]
    \item General Pick-and-Place Tasks: We evaluate \name{}’s performance when facing unseen objects, lighting variations, and novel scenes.
    
    \item Long-Horizon Manipulation Tasks: We measure the overall task success rate and the step-by-step completion accuracy of \name{} in extended manipulation sequences.
    
    \item Dexterous Dual-Arm Manipulation: We assess the model’s performance in tasks that require precise and coordinated control of both robotic arms.
\end{itemize}

To enable direct comparison with state-of-the-art models, we fine-tuned $\pi_0$ following the official instructions from its GitHub repository, using our self-collected data.

For the real-world general pick and place experiments, we collected 175 hours of pick-and-place data using our robotic platform and teleoperation interface. The dataset includes approximately 32,000 robot trajectories covering 30 different objects. We standardized the language command input as “put A into B”, where A represents the object to be picked and B denotes the target container.

For the baseline model $\pi_0$, we finetuned it directly on the robot trajectory data following its official training pipeline.
For \name{}, we adopted a different strategy — combining spatial-QA data with robot trajectory data, and leveraging both latent action labels and FAST labels during training.

\subsubsection{Genaral Pick and Place}

\begin{figure}[p]
    \centering
    \includegraphics[width=\linewidth]{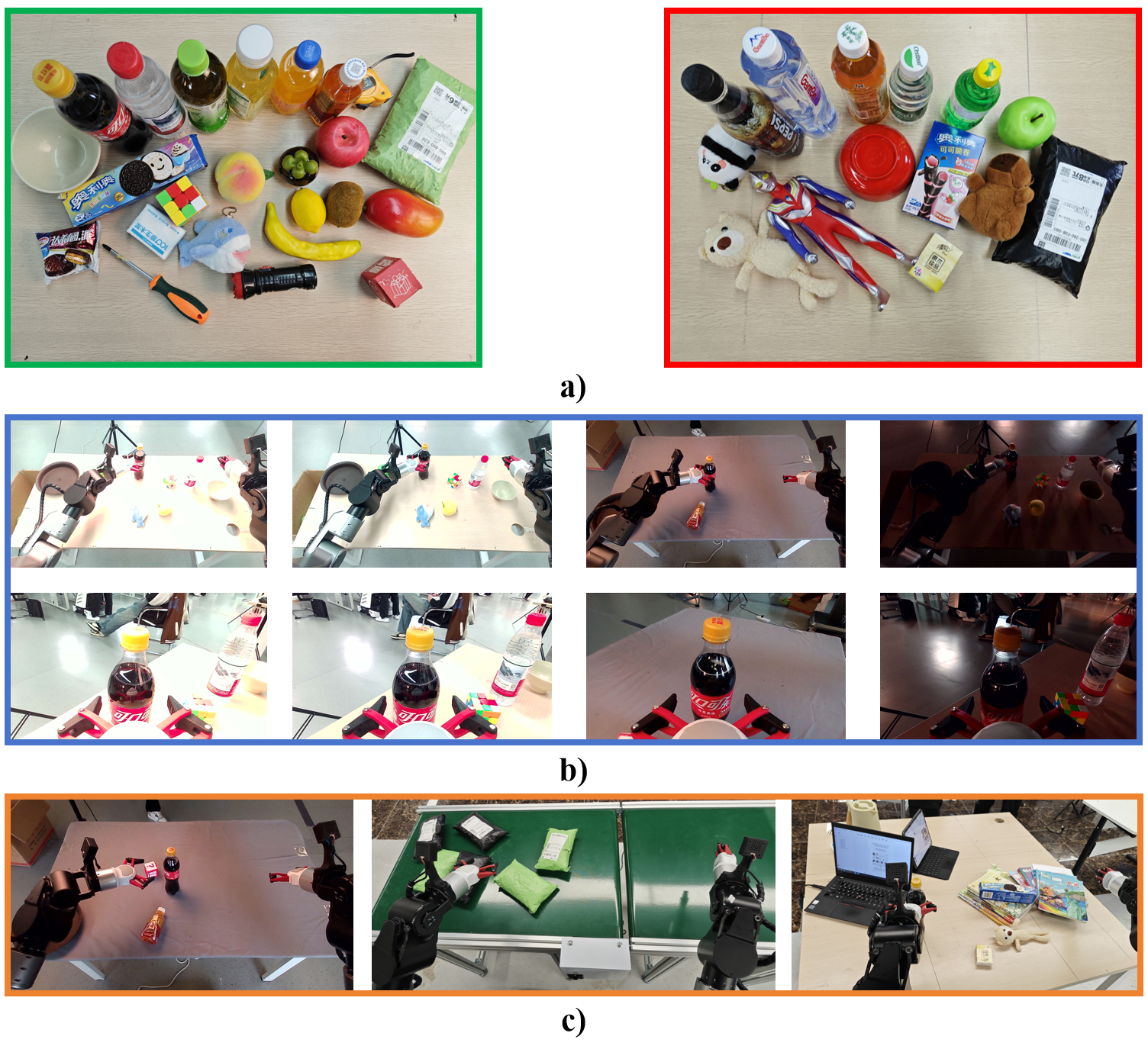}
    \caption{
        \textbf{Experiment Settings of Generalizable Pick-and-Place.}
        (a) Examples of seen and unseen objects — left: seen, right: unseen.
        (b) Illustration of varying illumination conditions.
        (c) Demonstration of grasping experiments in different scenes.
    }
    \label{fig:ppa-setting}
\end{figure}
Our evaluation was conducted under four different configurations. The \textbf{Basic} setting refers to test scenes and objects that also appear in the training data, with moderate and stable illumination. The \textbf{Unseen Objects} setting uses test objects that were not present in the training set but belong to the same semantic categories as seen objects, sharing similar noun descriptions in the language prompts. For example, an “Ultraman toy” not seen during training but belonging to the category “toy,” or an “Ice Red Tea” unseen during training but belonging to the category “beverage.” The detailed object categories are shown in Fig.~\ref{fig:ppa-setting}(a), where the left column lists seen objects and the right column lists unseen ones. The \textbf{Light Illumination Variation} setting keeps the same scenes and objects as the training set but introduces continuously changing light conditions, including extreme illumination levels not observed during training, as illustrated in Fig.~\ref{fig:ppa-setting}(b). The \textbf{Unseen Scenes} configuration tests the model on entirely different tabletop and environmental settings from those in the training set. While the training environment used square tables and cloth-covered surfaces, the test environments include alternative layouts such as factory conveyor setups and household scenes, as shown in Fig.~\ref{fig:ppa-setting}(c).

\begin{figure}[ht]
    \centering
    \includegraphics[width=\linewidth]{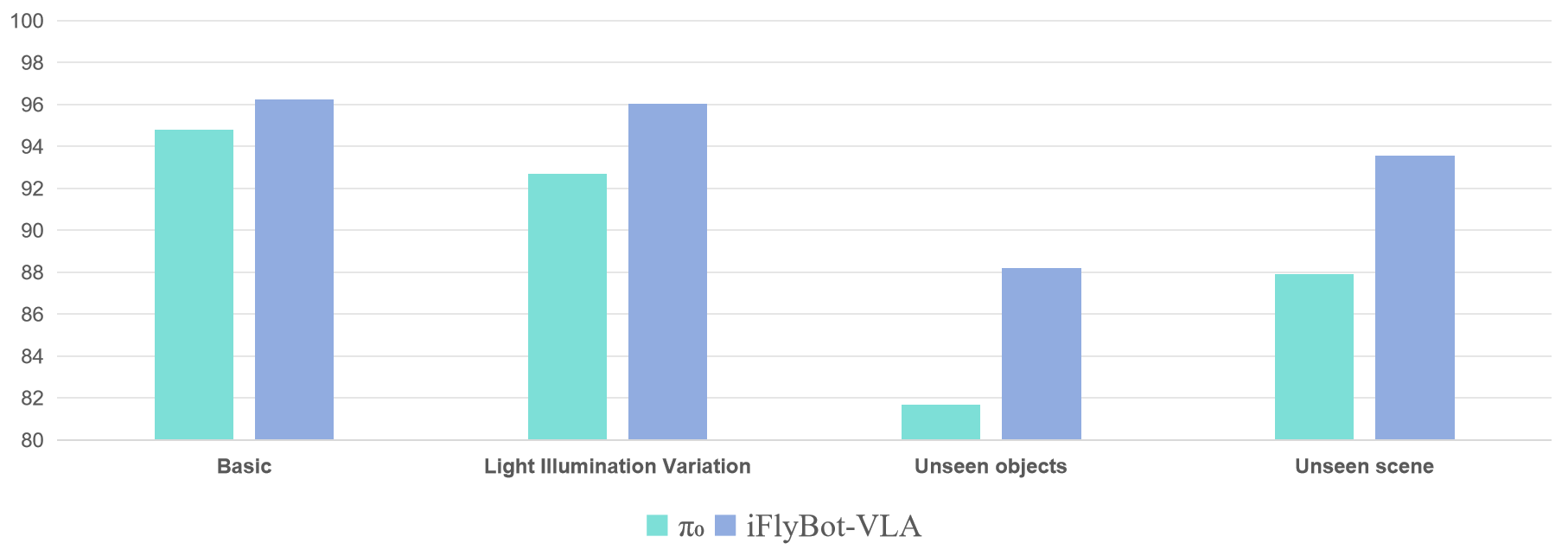}
    \caption{
        \textbf{Experiment Results of General Pick-and-Place.}
        Success rate of our policy and baseline.
    }
    \label{fig:ppa-result}
\end{figure}
For each configuration, we evaluated 24 seen object categories or 14 unseen object categories, performing 20 pick-and-place attempts for each. The results are presented in Fig.~\ref{fig:ppa-result}. 

From the experimental results, iFlyBot-VLA achieved success rates of \textbf{96.25\%}, \textbf{96.04\%}, \textbf{88.21\%}, and \textbf{93.57\%} in the Basic, Light Illumination Variation, Unseen Objects, and Unseen Scenes configurations, respectively. These results are slightly higher than those of the baseline model, which achieved \textbf{94.79\%}, \textbf{92.71\%}, \textbf{81.67\%}, and \textbf{87.91\%} under the same conditions. The improvements demonstrate that our proposed framework effectively enhances the model’s generalization capability across varying lighting, object, and environmental conditions.
\begin{figure}[t]
    \centering
    \includegraphics[width=\linewidth]{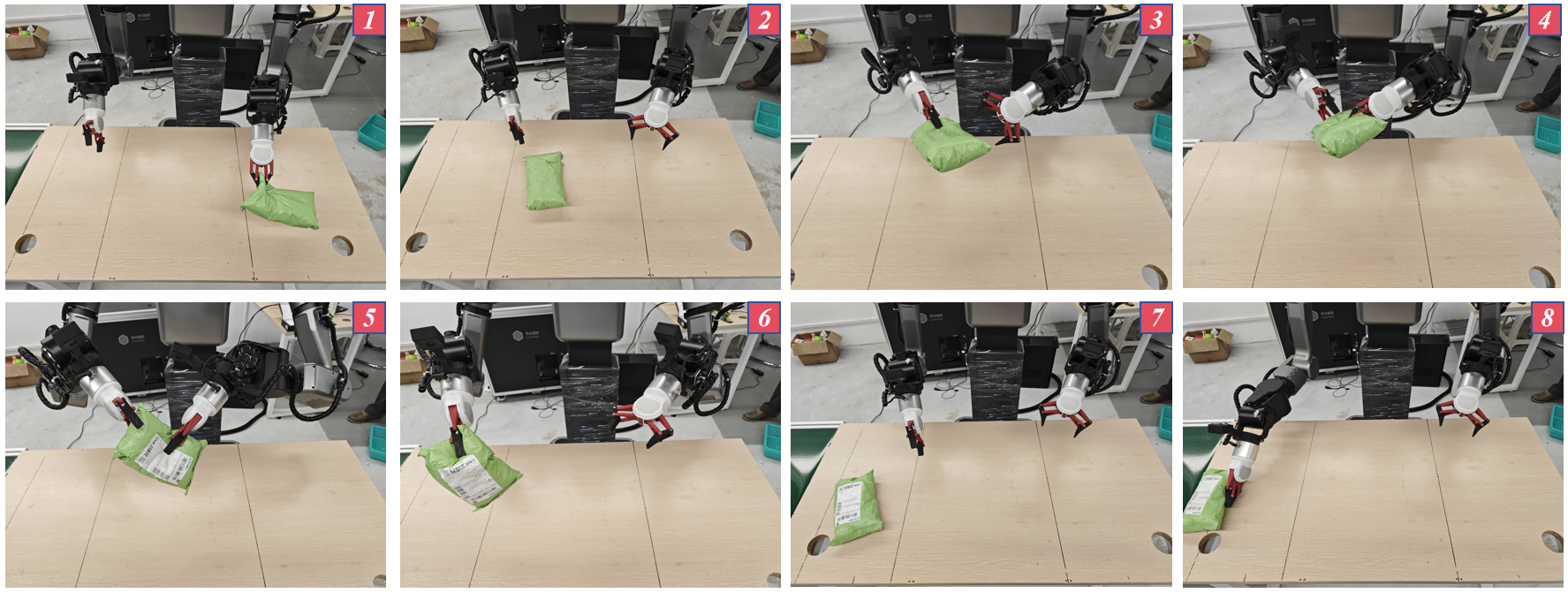}
    \caption{
        \textbf{Experiment Results of Long-Horizon Manipulation Tasks.}
        (a) Detailed steps of parcel sorting.
        (b) Results of our policy and baseline.
    }
    \label{fig:ppa-horizon}
\end{figure}

\subsubsection{Long-Horizon Manipulation Task}

\begin{figure}
    \centering
    \includegraphics[width=\linewidth]{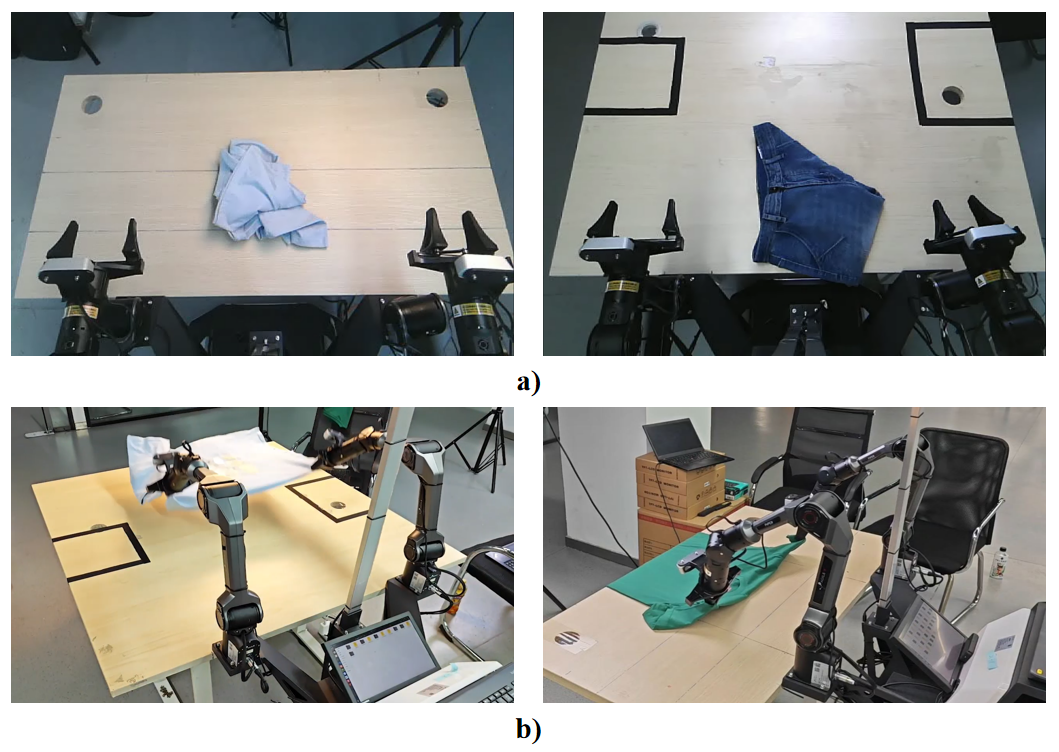}
    \caption{
        \textbf{Detailed Description of the Clothes Folding Task}
        (a) Initial state of the clothes.
        (b) Challenging actions involved in "flattening"
    }
    \label{fig:ppa-fold_task}
\end{figure}
We collected data for a “\textbf{parcel sorting}” task in a simulated factory conveyor-line environment, a long-horizon dual-arm manipulation task involving deformable packages, as illustrated in Fig.~\ref{fig:ppa-horizon}. To complete the full sorting process, the robot must: (1) grasp a flexible package, (2) determine whether the package orientation requires flipping, and if so, perform a coordinated dual-arm flipping motion, (3) place and push the package into the designated target area, and (4) repeat steps (1)–(3) until all packages on the table are sorted. Using the prompt \textit{“If the label is face-up, flip the package and put it in the basket”}, we collected approximately \textbf{47 hours} of data, corresponding to around \textbf{2,752 trajectories}.

Since the packages are soft and easily deformable, after the right arm places one corner of the package, its orientation may not be perfectly face-up. Therefore, we adopted two evaluation criteria: 
\begin{itemize}
    \item \textbf{Strict:} All packages must reach the correct target area, and their orientations must be correct immediately after the first placement by the right arm.
    \item \textbf{Allow Correction:} All packages must reach the correct target area, but if orientation errors occur after the right arm’s placement, up to two correction attempts are permitted.
\end{itemize}

We conducted 40 repeated trials, each involving three packages (two requiring flipping). The results show that under the \textbf{Allow Correction} criterion, iFlyBot-VLA achieved a \textbf{7.5\% higher success rate} than the baseline, highlighting the effectiveness of our dual-arm coordination and implicit planning strategies in handling deformable and long-horizon manipulation tasks.

\subsubsection{Challenging and Dexterous Dual-Arm Manipulation Task}

The \textbf{folding} task poses significant challenges for VLA models, as it requires both high-precision manipulation (e.g., grasping the correct corners of clothing) and robust policy control to handle the highly variable states of deformable objects. Previous studies have achieved success in folding already flattened clothes in real-world settings. However, when garments appear in arbitrary configurations—such as crumpled or twisted states—on the table, as shown in Fig.~\ref{fig:ppa-fold_task}(a), the task becomes substantially more difficult. 

\begin{figure}
    \centering
    \includegraphics[width=\linewidth]{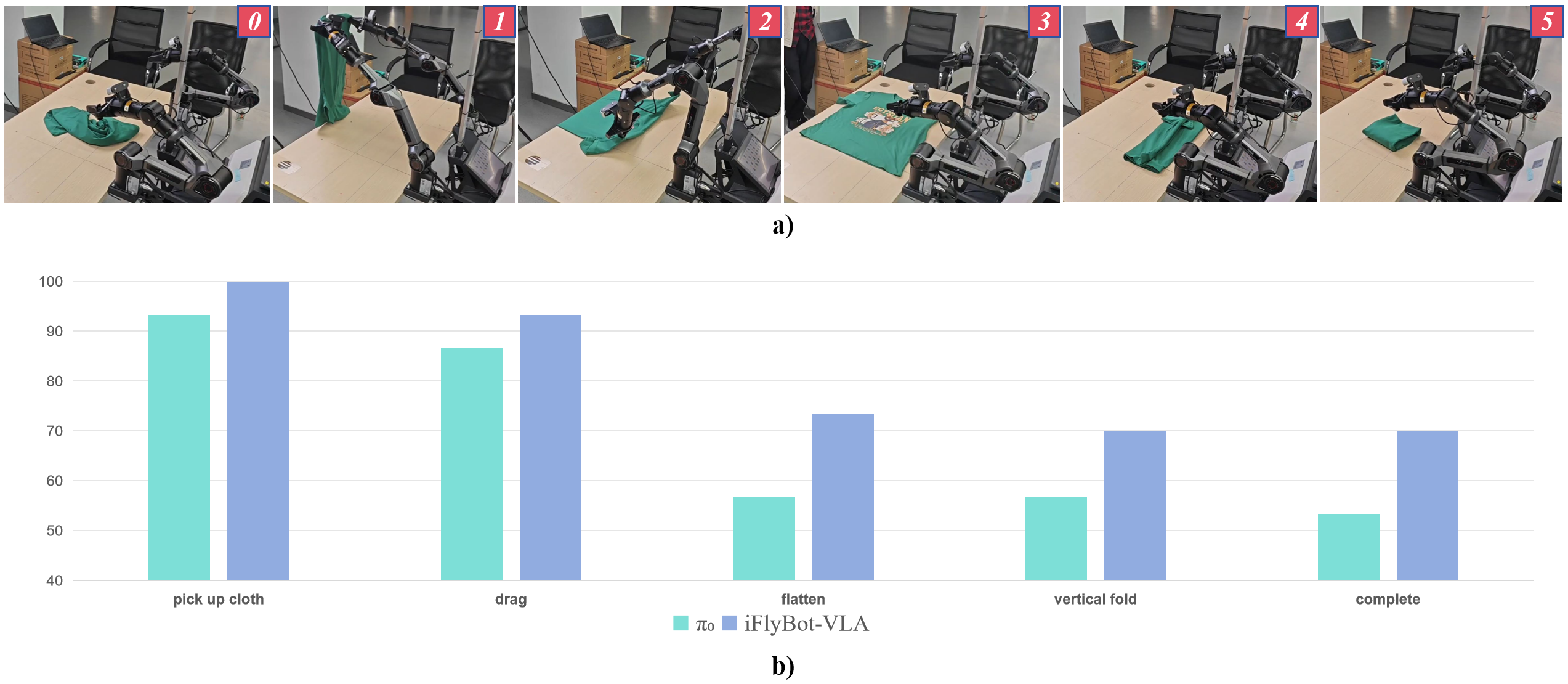}
    \caption{
        \textbf{We defined a step-by-step evaluation protocol and summarized the corresponding results as follows:}
        (a) Step Definition for the Folding Task.
        (b) Completion rate of each step.
    }
    \label{fig:ppa-fold_result}
\end{figure}
First, identifying the appropriate corner points to grasp from random configurations is highly challenging. Second, flattening the garment involves intricate sub-steps such as \textit{“flick flattening”}—which demands precise control over speed and acceleration, as shown in Fig.~\ref{fig:ppa-fold_task}(b) (left)—and \textit{“drag flattening”} along the table edge, which requires accurate perception of the garment’s state and key feature points, as shown in Fig.~\ref{fig:ppa-fold_task}(b)(right).

During testing, we observed that the success of the “\textbf{flick flattening}” motion depends heavily on the chunk-wise optimization strategy in the inference code. To ensure a fair comparison with baselines (e.g., $\pi_0$), we used the original inference code without modification and instead adopted the “\textbf{drag-flattening}” method. It is worth noting that although we did not employ the \textbf{flick-flattening} strategy, the combination of our model and the optimized inference policy achieved approximately 90\% success rate on this single step, as demonstrated in the video available on our project homepage. For this setup, we collected data on \textbf{8 clothing types} (5 T-shirts and 3 pairs of shorts), totaling approximately \textbf{110 hours} of demonstrations, with around \textbf{200 trajectories per clothing type}.

Given the complexity of this task, we report not only the overall success rate but also the success rate for each step to provide a more detailed comparison. Since locating the correct grasping points often requires multiple attempts, we imposed a \textbf{3-minute time limit} for each full execution. The detailed results are presented in Fig.~\ref{fig:ppa-fold_result}(b), where the x-axis corresponds to the steps illustrated in Fig.~\ref{fig:ppa-fold_result}(a). Note that since the flattening step may require multiple repetitions, when no overall time constraint is imposed, allowing sufficient time for corner-point searching and flattening enables \name{} to achieve nearly 90\% task success rate.


\section{Limitation and Conclusion}

\textbf{Limitation.} Although our model demonstrates outstanding performance in complex environments and long-horizon tasks, certain limitations remain. On one hand, despite the strong generalization capability of \textbf{iFlyBot-VLA} , it can still fail when following novel instructions involving unseen concepts or objects, and it faces challenges in grasping objects with shapes it has never encountered before. In future work, we plan to enhance the model’s ability to handle entirely new scenarios by scaling up the model, expanding the training dataset, and incorporating richer spatial representations.On the other hand, similar to all imitation learning approaches, \textbf{iFlyBot-VLA}  may struggle to maintain performance or recover effectively when encountering out-of-distribution inputs during inference. Therefore, in future research, we aim to integrate reinforcement learning (RL) mechanisms to further improve the model’s generalization ability and robustness in dexterous manipulation tasks, ultimately surpassing the inherent limitations of imitation learning and achieving stronger performance across more complex scenarios.

\textbf{Conclusions.} In this report, we introduced \textbf{iFlyBot-VLA} — a powerful \textbf{Vision-Language-Action (VLA)} model capable of generating action commands for controlling dual-arm mobile robots. We conducted an in-depth investigation of the model architecture and proposed a novel \textbf{explicit + implicit action planning framework}, which significantly improves model performance while introducing only a minimal increase in inference time.

Comprehensive experiments were conducted both in the \textbf{LIBERO simulator} and in real-world environments. The results demonstrate that iFlyBot-VLA excels in several key aspects: it generalizes effectively to unseen objects and environments, and exhibits remarkable robustness and reliability in executing both long-horizon and fine-grained manipulation tasks.

We believe that iFlyBot-VLA lays a solid foundation for the development of \textbf{general-purpose robotic systems} capable of assisting humans in a wide variety of everyday tasks, and that it contributes to advancing the field of general robotic intelligence.

\clearpage


\bibliographystyle{plainnat}
\bibliography{references}

\end{document}